\documentclass[sigconf]{acmart}
\AtBeginDocument{%
  }

\setcopyright{acmlicensed}
\copyrightyear{2026}
\acmYear{2026}
\acmDOI{XXXXXXX.XXXXXXX}
\acmConference[HRI '26]{ACM/IEEE International Conference on Human-Robot Interaction}{March 16–19, 2026}{Edinburgh, Scotland, UK}
\acmISBN{978-1-4503-XXXX-X/2026/06}

\usepackage{amsmath} 
\usepackage[ruled, lined, longend, linesnumbered]{algorithm2e}
\usepackage{xcolor}
\usepackage{graphicx}
\usepackage{enumitem}
\usepackage{makecell}
\usepackage{caption}
\usepackage{balance}
\usepackage[utf8]{inputenc}
\usepackage{mathtools}
\usepackage{amsmath}
\usepackage{tabularx,makecell,multirow,booktabs}
\newcolumntype{Y}{>{\centering\arraybackslash}X}

\SetAlgoNlRelativeSize{-1}           
\SetAlFnt{\small}                    
\SetAlCapFnt{\small}                 
\SetAlCapNameFnt{\small}             
\SetAlgoSkip{0.5em}                  
\DontPrintSemicolon   

\usepackage{xcolor}

\begin{document}

\title{Open-Ended Goal Inference through Actions and Language for Human-Robot Collaboration}

\author{Debasmita Ghose, Oz Gitelson, Marynel Vazquez, Brian Scassellati}
\email{debasmita.ghose@yale.edu}
\affiliation{%
  \institution{Yale University}
  \country{New Haven, CT, USA}
}

\renewcommand{\shortauthors}{Ghose et al.}
\begin{abstract}
  To collaborate with humans, robots must infer goals that are often ambiguous, difficult to articulate, or not drawn from a fixed set. Prior approaches restrict inference to a predefined goal set, rely only on observed actions, or depend exclusively on explicit instructions, making them brittle in real-world interactions. We present BALI (Bidirectional Action–Language Inference) for goal prediction, a method that integrates natural language preferences with observed human actions in a receding-horizon planning tree. BALI  combines language and action cues from the human, asks clarifying questions only when the expected information gain from the answer outweighs the cost of interruption, and selects supportive actions that align with inferred goals. We evaluate the approach in collaborative cooking tasks, where goals may be novel to the robot and unbounded. Compared to baselines, BALI yields more stable goal predictions and significantly fewer mistakes.
\end{abstract}

\ccsdesc[500]{Computing methodologies~Reasoning about belief and knowledge}

\keywords{Human Goal Prediction, Open-Ended Goal Discovery}


\begin{teaserfigure}
    \centering
    \includegraphics[width=0.85\linewidth]{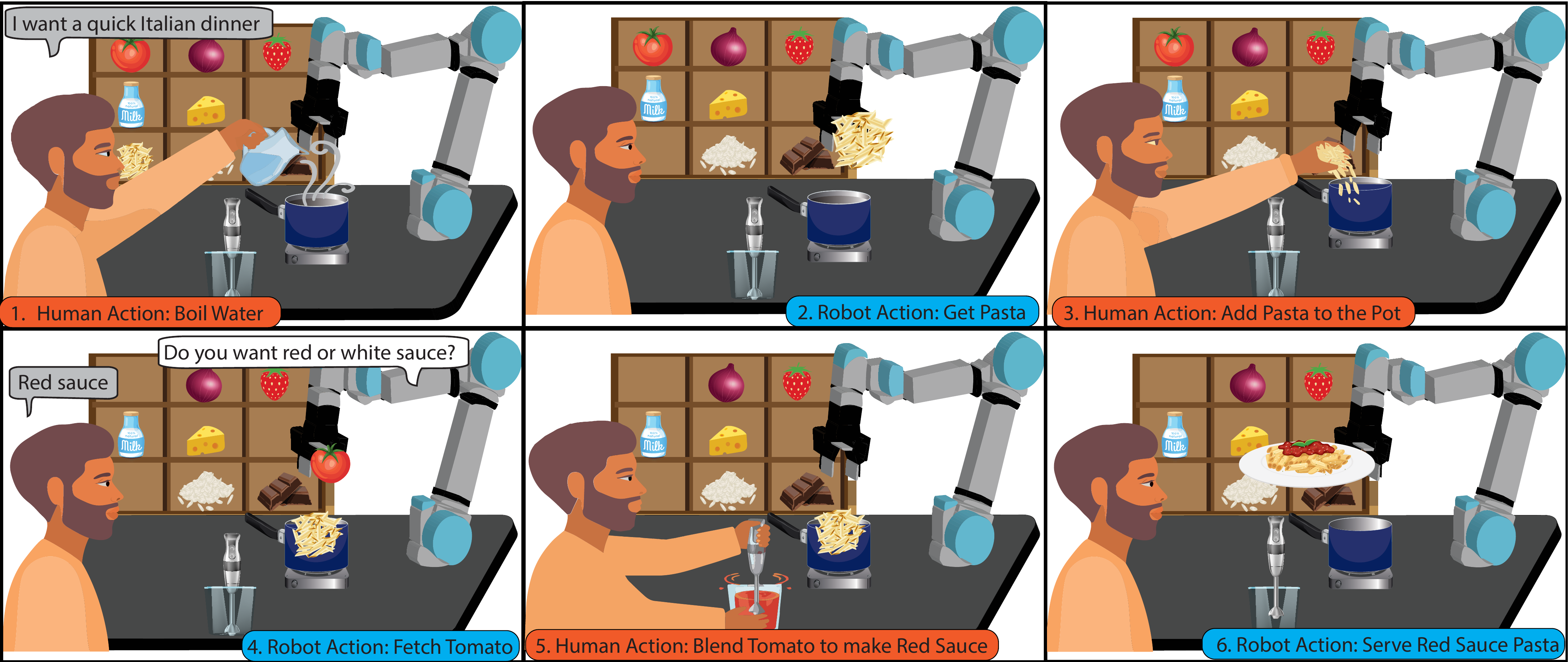}
    \caption{Illustrative cooking task. (1) Human says \textit{“I want a quick Italian dinner”} and boils water; the robot considers options that fit this preference (pasta vs. rice) and leans toward pasta. (2) Robot fetches pasta. (3) Human adds pasta to the pot, reinforcing a pasta goal. (4) Robot is uncertain about which pasta dish is intended since several variants are possible with the available ingredients, so it asks, \textit{“Do you want red or white sauce?”} Human replies \textit{“red,”} and the robot fetches a tomato as a result. (5) Human blends the tomato to make sauce. (6) Robot serves red-sauce pasta, completing the collaboration.}
    \Description{Placeholder teaser figure spanning both columns.}
    \label{fig:teaser}
  \end{teaserfigure}

\maketitle

\section{Introduction}



For robots to be effective partners in real-world human–robot collaboration, they must be able to anticipate and support the human’s end goals \cite{hoffman2024inferring}. Yet these goals are often unknown, difficult to articulate, or challenging for the robot to interpret, especially in long-horizon tasks where many intermediate actions overlap across different possible outcomes \cite{agrawal2022task}. Further, in many everyday settings, the space of potential goals is effectively unbounded: the same ingredients in a kitchen could be combined into countless recipes depending on preferences; the same tools and parts in an assembly task could produce different products; or a household reorganization could yield multiple equally plausible end states. To collaborate effectively in such settings, a robot must reason over this open-ended goal space while minimizing the burden on its partner, asking clarifying questions only when needed rather than demanding step-by-step instructions or demonstrations \cite{mannem2023exploring, unhelkar2020decision,liangintrospective}.

To address the above challenges, we propose  \textbf{\textit{Bidirectional Action Language Inference (BALI)}} for goal prediction, an approach for inferring human goals in collaborative tasks from both language and action. For example, consider a collaborative cooking task. When a person says \emph{“I want a sweet, healthy breakfast''} and then begins fetching oats and fruit, the robot must reason that the likely goals include preparing a \emph{fruit oatmeal} or a \emph{fruit parfait}. BALI leverages ambiguous verbal instructions and observed human actions to infer likely goals and guide robot behavior. Building on prior work, we propose to achieve this with Receding Horizon Planning (RHP), where the robot expands a tree of possible joint future actions toward plausible goals. But with open goal spaces, the tree can grow prohibitively large. Thus, BALI prunes unlikely branches by conditioning the tree expansion on the human’s instructions and observed actions, focusing on those most likely to support the human’s objectives. 
In effect, our approach enables the robot to use human actions to disambiguate the human's natural language description of their preferences. Similarly, it allows the robot to use explicit human preferences to guide its choice of supportive actions, establishing the bidirectional coupling between action and language.
Furthermore, when uncertainty about the human’s goal is high, such as distinguishing between \emph{fruit oatmeal} and a \emph{fruit parfait}, BALI allows the robot to selectively ask clarifying questions like \emph{“Do you want something warm or chilled?''}. BALI incorporates the human’s response to improve its inferences, as in Figure \ref{fig:teaser}. 

We evaluate BALI in simulated and proof-of-concept real-world collaborative cooking tasks. We compare against several approaches: using language alone, using actions alone without natural language instructions from the human, and using GOOD \cite{ma2024goal}, a recent approach that employs Bayesian reasoning for open-ended goal prediction using dialog. Our approach performs signficantly better than these approaches when goals are left open-ended. BALI makes fewer mistakes and maintains strong goal prediction accuracy while achieving higher accuracy. 
In sum, our main contributions are:
\begin{enumerate}[wide, itemsep=0.3em,labelindent=0em,topsep=0.3em]
    \item We introduce BALI, the first approach -- to our knowledge -- that combines natural language preferences and observed human actions for goal inference, where goal spaces can be open-ended.

\item BALI integrates a question-asking module that weighs goal uncertainty against interruption cost. Our experiments show that this  enables the robot to keep interventions minimal, asking clarifying questions only when needed, while facilitating goal inference.

\item We evaluate BALI in simulated and proof-of-concept real-world collaborative cooking tasks under open- and closed-world assumptions. In the open case, the goal set is unknown to the robot and potentially of infinite size. In the closed case,  the robot has access to the full set of human goals, which total 30 different options.  BALI  outperforms baselines with fewer mistakes, faster convergence, and higher goal-prediction accuracy.
\end{enumerate}

\section{Related Work}

In human-robot collaboration (HRC), a significant body of work addresses how robots infer human goals from observed actions (see Hoffman et al. \cite{hoffman2024inferring} for a survey). Most approaches assume a discrete hypothesis set and update a posterior over candidate goals via plan or intent recognition. This includes goal inference in shared workspaces \cite{hoffman2007cost, el2022hierarchical, adamson2021we}, in shared-autonomy \cite{jonnavittula2021know, qiao2021learning}, and navigation with fixed destinations \cite{bandyopadhyay2013intention, kollmitz2015time}. In these settings, inference is typically short-horizon, estimating instantaneous intent in shared autonomy from recent inputs \cite{jain2019probabilistic, aronson2021inferring, aronson2022gaze, decastrodreaming}, predicting the next step in workspace collaboration \cite{nikolaidis2013human, haninger2022model, tung2024workspace}, or forecasting navigation sub-goals \cite{thompson2024predicting}. Extensions expand the temporal scope through hierarchical or dynamic models and active disambiguation, though still within finite goal sets, as in Bayesian Delegation \cite{wu2021too}, informative control for self-driving \cite{sadigh2016information}, and workspace reconfiguration \cite{tung2024workspace}. Other work explores alternative inference mechanisms over fixed goal banks, such as framing goal recognition as reinforcement learning \cite{amado2022goal}, combining action sequences with spoken language via Bayesian inverse planning with LLM-based likelihoods \cite{ying2023inferring}, or augmenting plan recognition with dialog to refine beliefs \cite{idrees2023improved}.

A complementary line of research investigates intrinsic motivation, where robots autonomously generate and pursue open-ended goals. Prior work exploits intrinsic signals such as curiosity or uncertainty reduction \cite{qureshi2018intrinsically, romero2025motivational}, or relies on demonstrations and tutoring to acquire parameterized policies for parameterized goals \cite{nguyen2012active, nguyen2014socially, thomaz2007robot, sao2024intrinsic}. While this work targets lifelong goal discovery rather than inference of a collaborator’s current task goal, it motivates open-ended goal reasoning.

More recently, several approaches have inferred goals without a predefined goal set. Zhi et al. \cite{zhi2024infinite} cast goal inference as a Particle Filter over open-ended goal spaces, proposing goals from subgoal statistics and reweighting via inverse planning on observed actions. Ma et al. introduce GOOD \cite{ma2024goal}, which maintains a posterior over unbounded natural-language goals using an LLM to score dialog. Our approach lies at this intersection: unlike methods that rely only on action traces \cite{zhi2024infinite} or only on dialog \cite{ma2024goal}, we jointly reason over both observed actions and natural-language signals. This enables maintaining an open-ended goal posterior, predicting a human’s \emph{long-term} goal, and asking clarifying questions when the expected information gain outweighs the interruption cost.



\section{Preliminaries}
\label{sec:preliminaries}

\noindent
\textbf{Problem Formulation.}
A robot aims to support a human in a collaboration, where they take action one after the other. Let the environment's state be $s$, and a shared action space be $\mathcal{A}$. We assume that the state and the action spaces are fully observable. 
The action space contains high-level actions (e.g., in a cooking task, the action space can include boiling water, chopping vegetables, etc.). Both agents have access to a shared action history $\mathcal{H}^{0:t-1}$ up to (but not including) the current time step. We also keep track of the individual action histories of the human and the robot, denoted by $\mathcal{H}^{0:t-1}_h$ and $\mathcal{H}^{0:t-1}_r$. Further, 
the human expresses a set of preferences for the task in natural language ($\mathcal L = l_1 \dots l_n$) that can be open-ended and ambiguous, like \emph{"I want a savory dinner meal.''} 

The robot seeks to reduce uncertainty about the human's current goal so that it can take actions that support them. At each timestep, the robot infers $p(g)$, a probability distribution over the human's possible goals $\mathcal{G}$, and at alternate timesteps, selects a robot action $a_r$. The set of all possible human goals $\mathcal{G}$ is potentially unbounded, so we propose for the robot to continuously estimate which goals are plausible given observations of the human's actions and speech. 

\vspace{0.5em}
\noindent
\textbf{Receding Horizon Planning (RHP).} 
Following ~\cite{ghose2024planning, sadigh2016information}, the robot chooses actions using  Receding Horizon Planning (RHP)~\cite{ma2006receding}.
Model Predictive Control (MPC) inspires RHP \cite{garcia1989model}, but RHP applies to discrete decision-making rather than continuous control. At each timestep, the robot expands a search tree, 
where each node is a state $s$ over a finite planning horizon. It evaluates possible sequences of human and robot actions conditioned on the robot’s current belief about the goal. Then, the robot executes the first action of the sequence that maximizes expected support for the human’s goal. Finally, the horizon window shifts forward, the robot updates its belief over goals using the newly observed human action, and planning repeats. As explained later in Sec.~\ref{sec:BALI}, BALI expands this formulation to also account for verbal human preferences.

\vspace{0.5em}
\noindent
\textbf{Attractor Fields.} 
Per Jain and Argall \cite{recursive_bayesian},  \textit{an attractor field expresses the extent to which one element in the first set exerts a pull toward certain elements in the second set. }
Formally, let a target set $Y$ have $n$ elements (e.g., $n$ actions). An attractor field for a source element $x$ (e.g., a goal) is a vector:
\begin{equation}
    F_x = \big[F_x(1), F_x(2), \dots, F_x(n)\big], \quad F_x(i) \geq 0,
    \label{eq:attractor}
\end{equation}
where $F_x(i)$ indicates how strongly $x$ "pulls" on the $i$-th element in $Y$. If multiple sources are present, their fields are summed element-wise. As a result, targets that are relevant to more sources accumulate higher attraction scores, while those relevant to fewer sources accumulate weaker scores. For example, in a collaborative cooking task, a goal such as \textit{``fruit oatmeal''} exerts an attractor field over actions like \textit{fetching oats, cutting fruit}, and \textit{boiling water}. A high-level preference like \textit{"healthy''} can also be modeled as an attractor field, pulling the robot toward ingredients such as \textit{fruit} and away from high-sugar toppings like \textit{chocolate chips}. Similarly, appliances can act as attractors: a \textit{blender} exerts a strong pull toward goals like \textit{“smoothie''} but little or no pull toward goals like \textit{“salad.''}

Attractor fields can be computed using known statistical mappings between the source and target elements or the world knowledge captured by large language models (LLMs). More specifically, given a source element (e.g., a goal like \textit{"fruit parfait''} or a preference like \textit{“sweet''}), we propose to use an LLM-as-a-judge \cite{gu2024survey} to score the relevance of possible targets (e.g., assigning higher scores to \emph{``cut fruit''} than to \emph{``fry bacon''}, or to \emph{``blueberries''} rather than \emph{``lettuce''}). We use these scores to populate the attractor field vector, allowing the robot to construct attractor fields even when the source is open-ended, ambiguous, or novel.


\begin{algorithm}[t]
\caption{BALI Summary}
\let\oldnl\nl
\newcommand{\nonl}{\renewcommand{\nl}{\let\nl\oldnl}}

\label{alg:bali-summary}
\small
\KwIn{state $s$, action space $\mathcal{A}$, language pref. $\mathcal{L}$, horizon $H$, prior belief over goals $p(g)$, human history $\mathcal{H}_h$, robot history $\mathcal{H}_r$}
\KwOut{next robot action $a_r$, updated belief over goals $p(g)$}

  Observe human action; update $\mathcal{H}_h$\;
  Generate interaction summary\;
  Predict plausible human goals ($g_{pred}$) and goal uncertainty $p(g)$\;
  \If{goal uncertainty is high and interruption cost is low}{
    Ask clarifying question; update $p(g)$ and attractor fields\;
    $\mathcal L$ = $\mathcal L$ + human's answer to question\;
  }

  Initialize RHP tree with root $s$\;
  \For{depth $=1$ \KwTo $H$}{
    \ForEach{node at this depth}{
      Enumerate valid robot actions $\mathcal{A}_{\text{valid}}$\;
      Score each $a \in \mathcal{A}_{\text{valid}}$ using goal- and preference-based attractor fields\;
      Filter to most likely actions based on scores\;
      Expand node with filtered actions\;
    }
  }

  Evaluate branch costs; select minimum-cost branch; execute its first action $a_r$; update belief over goals $p(g)$\;
  \nonl{\textbf{Return:}} next robot action ($a_r$), updated belief over goals ($p(g)$)

\end{algorithm}

\begin{figure*}
    \centering
    \includegraphics[ width=0.88\linewidth]{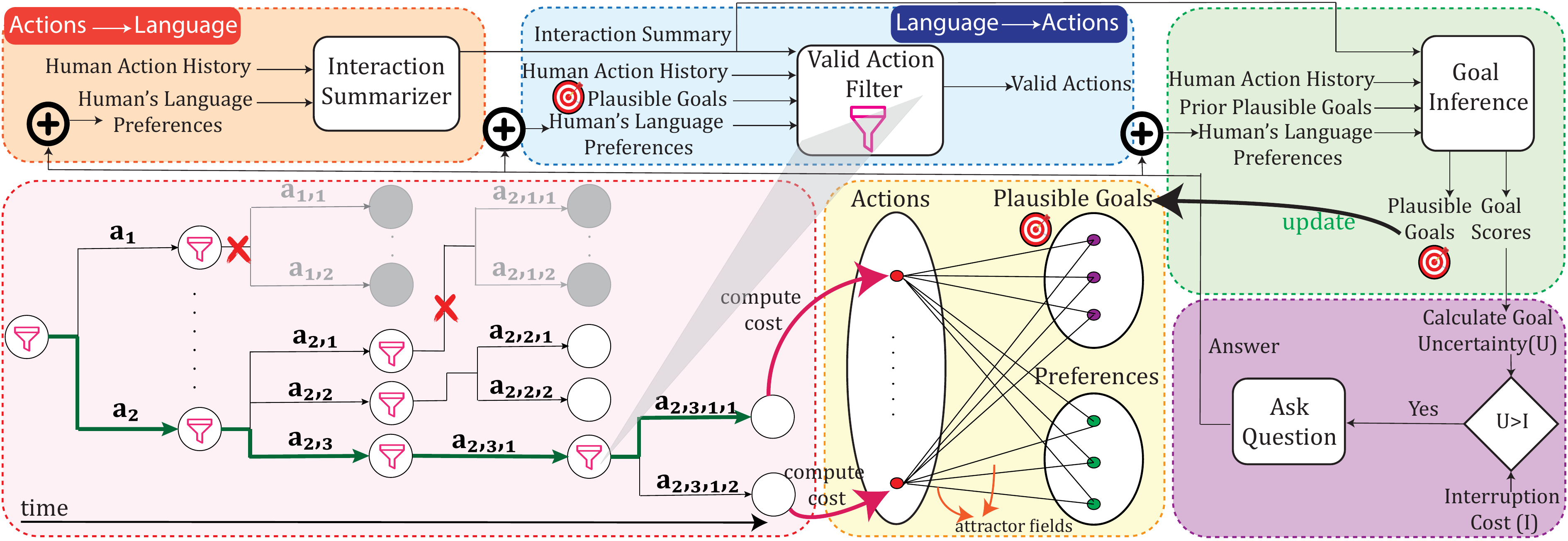}
    \caption{BALI for goal prediction: Human preferences and actions are summarized (orange) and used for Goal Inference (green) to update an estimate of plausible goals. The Ask Question module (purple) triggers clarifications when goal uncertainty exceeds a threshold. The planner (pink) expands action sequences under receding-horizon planning: at each non-leaf node, the Valid Action Filter (blue) prunes infeasible actions (shown greyed out). The cost function module (yellow) computes attractor field cost linking actions to goals and guiding the search toward trajectories aligned with plausible human goals and preferences.}
    \label{fig:system}
    \vspace{-1em}
\end{figure*}

\section{Bidirectional Action-Language Inference (BALI) for Goal Prediction}
\label{sec:BALI}

Our approach, BALI, is designed for turn-based human-robot collaborations (as explained in Sec.~\ref{sec:preliminaries}) and 
%
is summarized in Algorithm~\ref{alg:bali-summary}. 
At each time step, the robot observes the human’s latest action and updates its belief over possible goals using attractor fields that link actions to preferences expressed in natural language and plausible goals. If uncertainty is high and the cost of interruption is low, the robot asks a clarifying question, incorporating the human’s response to refine its beliefs (lines 4-7). The robot then performs receding horizon planning (lines 9-16): it expands a search tree of alternating human and robot actions up to a fixed horizon ($H$), where at each node valid robot actions are enumerated, scored with respect to goals and preferences, filtered to retain the most likely, and expanded. The resulting tree is evaluated, and the branch with the minimum cost is identified. The robot executes the first action along this branch as its next contribution to the task (line 17). The rest of the section explains each key module in BALI in more detail.

\subsection{Interaction Summary}
\label{sec:interaction_summary}

We maintain a concise interaction summary that condenses a human’s preferences and actions into a single natural-language sentence, inspired by the concept of Dialog State Tracking (DST) in conversational systems \cite{jacqmin2022you}. For example, in a cooking task, the summary describes the likely dish, the current phase (e.g., gathering, assembling, finishing), the items in play, and sensible next steps (orange box in Fig. \ref{fig:system}). We encode the history in natural language because our goal inference leverages large language models, which inherently reason over text-based representations. The summary provides the planner with a consistent reference across timesteps, preventing the robot from overreacting to isolated ambiguous actions, enforcing coherence as the interaction unfolds, and preserving long-range context. Moreover, this module helps the robot use human actions to disambiguate and ground high-level instructions.

\subsection{Goal Inference}
\label{sec:goal_inference}
The Goal Inference module (green box in Fig. \ref{fig:system}) enables a robot to infer goals by analyzing the action history and updating a ranked list of candidate goals ($g_{pred}$) as new actions are taken.
Conditioned on the human’s preferences, action history, the interaction summary, and environmental constraints, the robot queries an LLM that leverages commonsense knowledge
to generate and continually update a set of plausible goals, inspired by Ma et al. \cite{ma2024goal}. Each goal is also assigned a score, using LLM-as-a-judge, that indicates how aligned the goal is with the human's action history and preferences.  
The system handles removing and adding goals by continuously re-evaluating the candidate set as new human actions and preferences are observed. When new actions or preferences indicate that specific goals are no longer supported, they are removed. Conversely, they are added if the observed behavior 
support new feasible goals. 

\subsection{Action Selection and Cost Function}
\label{sec:action_selection}

BALI chooses robot actions within the receding horizon planning (RHP) framework described in Sec.~\ref{sec:preliminaries} (pink box in Fig.~\ref{fig:system}). Action selection proceeds by expanding a search tree of possible future action sequences, alternating between human and robot actions, up to a fixed planning horizon $H$. At each node, a valid action filter (blue box in Fig.~\ref{fig:system}) uses the interaction summary (Sec.~\ref{sec:interaction_summary}), goal predictions (Sec.~\ref{sec:goal_inference}), and action history to select actions consistent with the current goals and preferences ($\mathcal A_{valid}$) using an LLM. The filtering repeats at each level of the RHP tree until depth $H$, generating candidate action sequences. After the sequences are generated, attractor fields are computed per Eq.~\ref{eq:attractor} (yellow box in Fig.~\ref{fig:system}), producing relevance scores for each candidate action with respect to every plausible goal ($\forall g \in g_{\text{pred}}$) and preference ($\forall l \in \mathcal{L}$). 
The robot aggregates the attraction scores from all plausible goals and preferences to assign a single scalar value to each candidate action along every branch of the RHP tree. Formally, for each $a \in \mathcal{A}_{valid}$, the aggregate attractor score is:
%
\begin{align}
{\small
    S(a) \;=\; \sum_{g \in g_{\text{pred}}} w_g\, F_g(a) \;+\; \sum_{l \in \mathcal{L}} w_l\, F_l(a),
}
    \label{eq:S}
\end{align}
\vspace{-0.05em}
\noindent
where $F_g(a)$ and $F_l(a)$ are the attractor field scores for goals ($g$) and language preferences ($l$), and $w_g$, $w_l$ are tunable weights reflecting the importance of attractor fields for $g$ and $l$, respectively over $a$. 

The cost of a branch in the RHP tree, which encodes an action sequence $\mathbf{a}=(a_1,\dots,a_H)$ for the human and robot, is $C(\mathbf{a}) \;=\; -\sum_{t=1}^H S(a_t)$ with $S$ as in eq.~(\ref{eq:S}). The robot identifies the branch with minimum cost (the green line in the pink box of Fig.~\ref{fig:system}), and executes the first action ($a_r$) along that branch ($a_2$ in Fig.~\ref{fig:system}).

\subsection{Asking Informative Questions}
\label{sec:questions}

The robot must decide \textbf{\emph{when}} to ask a question, \textbf{\emph{what}} question to ask and \textbf{\emph{how}} to use the answer to reduce uncertainty over goals, as shown in the purple box in 
Fig. \ref{fig:system}. 

\subsubsection{When to Ask}

The robot asks a question only if the expected benefit (information gained about the human’s goal) outweighs the cost of interrupting. Inspired by \cite{unhelkar2020decision}, we model the cost as time-dependent to discourage repeated interruptions: immediately after asking, the interruption cost is high and then decreases linearly toward a minimum. Let $\Delta t$ be the time since the last question, $T_q$ the interval after which the minimum cost is reached, and $\Delta C = C_{\max} - C_{\min}$ the range of costs. The cost of asking at time $t$ is:  

\begin{align}
\vspace{-1em}
{\small
C_q(t) =
\begin{cases}
C_{\min}, & \Delta t \ge T_q, \\[6pt]
C_{\max} - \Delta C \cdot (\Delta t/T_q), & \Delta t < T_q,
\end{cases}
}
\end{align}  

The robot compares this interruption cost to its current uncertainty or entropy over goals $p(g)$: $H(g) = -\sum_g p(g)\log p(g).$
Since the value of resolving uncertainty increases with the number of plausible goals ($n_{goals}$), we scale the interruption cost by $\log(n_{\text{goals}})$: 
$C_q^{\text{scaled}} = C_q(t)\,\log(n_{\text{goals}}).$
A question is asked whenever $H(g) > C_q^{\text{scaled}}$, i.e., when the robot’s uncertainty is high enough that reducing it is worth the interruption.  

\subsubsection{What to Ask}  

Once the robot decides to ask, it must choose the most informative question. Candidate questions are drawn from categories relevant to the task (e.g., in cooking: high-level food preferences, ingredients, or appliances). These categories can be predefined or dynamically generated by an LLM based on the domain. Each candidate question is then hypothesized to have several possible answers from the human.

The usefulness of a question is measured by how much it reduces uncertainty about the human’s goal. We quantify uncertainty by the entropy of the current belief distribution $p(g)$ over goals:
$H = -\sum_g p(g)\log p(g).$ Inspired by \cite{biyik2019asking, fitzgerald2022inquire, qian2024pps}, when the robot asks a question, each possible answer ($ans$) changes its belief over goals, yielding $p(g \mid ans)$ whose entropy reflects how uncertain the robot would be if that answer were given. We compute this posterior probability $p(g \mid ans)$ using Bayes' rule: $p(g\mid ans)=\dfrac{p(ans\mid g)p(g)}{p(ans)}$.
The likelihood of each answer given a goal, $p(ans\mid g)$, is approximated using attractor fields, which encode how strongly each answer is associated with the goal. For example, if the question is \emph{“Should the dish be warm or chilled?”}, then goals like \emph{``oatmeal''} have high attraction to the \emph{“warm”} answer and low attraction to the \emph{“chilled”} answer, while \emph{``parfait''} has the reverse pattern. $p(g)$ denotes the probability distribution over possible human goals estimated by the Goal Inference module (Sec.~\ref{sec:goal_inference}). For simplicity, we assume that each possible human answer is equally likely, i.e., $p(ans)$ is uniform.
Therefore, the expected uncertainty after asking is the weighted average entropy for all possible answers: $ \mathbb{E}[H \mid ans] = \sum_{ans} p(ans)H(p(g \mid ans))$, and the question's value is its expected reduction in entropy: $ \Delta H = H - \mathbb{E}[H \mid ans].$


Intuitively, a good question partitions the plausible goals into groups depending on the answer, so that the robot’s uncertainty decreases regardless of how the human responds. For instance, like the earlier example, asking \emph{“Should the dish be warm or chilled?”} separates \emph{``oatmeal''} from \emph{``parfait''} and removes ambiguity.

\subsubsection{Processing Human Answers}  
Human answers are incorporated into BALI in three ways (see the $\oplus$ symbol in Fig. \ref{fig:system}):  
(i) the Interaction Summary module (Sec.~\ref{sec:interaction_summary}) integrates the answers into the natural-language summary of preferences and actions,  
(ii) the Goal Inference module (Sec.\ref{sec:goal_inference}) uses the answers as additional evidence when predicting the human’s goal, and  
(iii) the Action Selection module (Sec.~\ref{sec:action_selection}) conditions the RHP tree expansion on the updated goals, producing actions that align with new preferences.

\section{Evaluation}

\subsection{Task and Problem Representation}

We evaluate BALI on collaborative cooking, a testbed widely used in human–robot collaboration \cite{brawer2023interactive, carroll2019utility, van2022correct, goubard2023cooking}. Cooking provides concrete ground truth for controlled experiments while also reflecting the subjective and often unique preferences that make goals both explicitly nameable and inherently emergent. Our experiments span both a simulation and a real-world setup. 
The human can prepare any meal using the available ingredients and appliances. 
We assume either partner may perform any preparatory step, with each step taking the same time regardless of who executes it. 

The robot’s state and action space are defined in the Planning Domain Definition Language (PDDL) \cite{aeronautiques1998pddl}, with environment updates handled by PDDLgym \cite{silver2020pddlgym}. Both agents can perform actions \begin{math}
    A = \{\texttt{gather(i,a)}, \texttt{pour(i,d,a)},  
    \texttt{mix(i,a)}, \texttt{cook(i,a)},
    \texttt{turn\_on(i,a)},\\
    \texttt{collect\_water(a)}, 
       \texttt{blend(i,a)},
       \texttt{reduce\_heat(i,a)}, 
      \texttt{serve(i,a)}\}
\end{math}
where $i$ is an item, $d$ a container (e.g., a bowl), and $a \in \{human, robot\}$ denotes the acting agent. The state is represented as a vector of PDDL literals describing item locations and states (simmering, blended, cooked). We assume a deterministic, fully observable environment. For simplicity, the human initiates the interaction by taking the first action and providing high-level meal preferences (e.g., \textit{“I want something sweet and healthy”}).

\subsection{Experimental Setup}

We primarily assess BALI in an open-ended goal inference setting, which we refer to as the \textbf{\textit{open case}} going forward. In this setting, we conduct experiments both in simulation and in the real-world. Additionally, for completeness, we investigate performance in simulation in a \textit{\textbf{closed case}} where the goal set is known. This closed setting includes additional baselines not possible in the open case.

\subsubsection{Simulation}

The human prepares one of 30 possible goal recipes, where each recipe could be one of six types: \textit{Pasta, Stew, Salad, Oatmeal, Smoothie,} or \textit{Parfait}, in both the open and closed cases.
The human’s ground-truth action sequences are derived from Clique/Chain Hierarchical Task Networks  (CC-HTNs) \cite{hayes2016autonomously}, capturing ordered steps (e.g., add strawberry before turning on the blender) and steps that can be performed unordered (e.g., retrieve stew toppings in any order). 
To construct the simulated preferences, we used GPT-o3 to propose 50 diverse food preferences given the available ingredients, appliances, and exemplar recipes (the 30 goal recipes the simulated human could be preparing).
For example, the preferences include \emph{``sweet", ``savory", ``warm", ``breakfast",} etc.  We manually mapped each of the 30 recipes to all applicable preferences. Next, we generated 967 preference pairs by taking all unordered pairs of preferences that shared at least one goal. For example, the overlap of ``\emph{sweet''} and ``\emph{breakfast''} includes ``\emph{fruit parfait''}. From each pair, we uniformly sampled one overlapping goal to execute, resulting in 967 experiments. The length of simulated human recipes ranges from 9 to 31 steps, with an average length of $22.7 \pm 5.5$ steps.



\subsubsection{Real World}

Similar to Fig.~\ref{fig:teaser}, our real-world setup comprises a UR-5e, an overhead Microsoft Azure Kinect \cite{smisek20133d} for workspace monitoring, and a microphone. To avoid online action recognition, humans were instructed to verbalize their actions. A custom lightweight parser used fuzzy matching to map utterances to PDDL literals. For safety, we used realistic mock ingredients (liquids replaced with colored beads) and modified appliances to actuate without heat or blades. A blender and a stove with a pan were placed in front of the robot. A sink, utensils, and condiments were to the left,  serving vessels were to the right, and a storage shelf with ingredients was behind. 
For perception, we used SEEM \cite{zou2023segment}, a promptable segmentation model, to localize and segment ingredients in different colored containers from text prompts. Then, we applied a heuristic grasp-point selector on the resulting mask to enable manipulation.

\subsubsection{Implementation Details}

When comparing approaches that rely on LLM models, we use a \emph{larger} general-purpose LLM (Gemini 2.5 Flash \cite{comanici2025gemini}) and a \emph{smaller} instruction-tuned model (Phi-4-mini-instruct \cite{abouelenin2025phi}), with identical prompts and decoding settings. Furthermore, when our method asks questions, it uses the clarifying question module introduced in Sec. \ref{sec:questions}. In the open case, we use the FlowJudge \cite{flowaiFlowJudge} model as the LLM-as-a-judge for attractor field scoring. In the closed case, we leverage information about the goals as explained further in the next section.

\subsection{Baselines}
\label{sec: baselines}

We consider five baselines in our evaluation. Two baselines (LLM Only, and GOOD) are completely different methods to BALI. The other three have some similar components to BALI, but they either use less information from the human, or leverage access to a goal bank (which is generally not available to BALI in the open case).


\subsubsection{LLM Only}This baseline examines a purely language-driven approach: Can a large language model, given a user’s stated preferences and the valid actions at each step, select sensible next actions for the robot without explicit planning or an internal belief over goals? At each timestep, the system lists the valid robot actions from the current state and prompts an LLM with the user’s active preferences (and brief interaction context) plus the valid action list. The model is instructed to return exactly one action from that list that best supports the human's preferences, leveraging the model's general cooking knowledge. In the \textbf{\textit{open case}}, the prompt contains only preferences and the action list. In the \textbf{\textit{closed case}}, it additionally includes the set of possible goals (the 30 recipes) so the model can use this information to guide its action choice. Decisions are made one step at a time, without considering possible futures.

\subsubsection{Goal Inference from Open-Ended Dialog (GOOD)} Goal Inference from Open-Ended Dialog (GOOD) \cite{ma2024goal} maintains a posterior over open-ended, natural-language goals that is updated from dialog alone. At each turn, it prompts an LLM to role-play the user under each candidate goal, uses the resulting likelihoods to perform Bayesian updates, and adds/prunes goal hypotheses as the conversation evolves. In our setup, we run this dialog→inference→goal-management loop and map the current most probable goal to a next action from the valid action set. In the \textbf{\textit{open case}}, GOOD maintains a posterior over unbounded natural-language goals. In the \textbf{\textit{closed case}}, the posterior is restricted to the known recipes. 
Unlike our method, GOOD neither looks ahead to predict future action sequences nor conditions action selection on past human actions.

\subsubsection{Actions Only}This baseline aims to evaluate goal inference based on the human’s action history, ignoring natural-language preferences. 
In the \textbf{\textit{closed case}}, we implement this baseline using Critical Decision Points (CDP) \cite{ghose2024planning}, as this approach uses RHP and infers a human's long-horizon goals considering past human actions. The CDP approach identifies states where candidate goals diverge most in their next-step predictions, and chooses the robot action that steers the interaction toward such informative states. Because CDP relies on a goal-driven policy bank, 
in the \textbf{\textit{open case}} we expand an RHP tree, but use a language model to score admissible next actions based on perplexity \cite{ankner2024perplexed} given the current interaction sequence, treating lower perplexity as more likely. In this context, perplexity measures how surprising the model finds future actions in the sequence; actions that yield lower perplexity are considered more consistent with the interaction so far. We keep the top-K actions, look a few steps ahead with the model to score candidate paths, and execute the first action from the best-scoring path.


\subsubsection{Known Goals (BALI-KG): }
This baseline is designed only for the \textit{\textbf{closed case}}. We apply our method but restrict the Goal Management module to a fixed goal bank $\mathcal{G}$ of known recipes. All other components—interaction summarization (Sec.~\ref{sec:interaction_summary}), RHP tree expansion, and attractor-field cost computation (Sec.~\ref{sec:action_selection}) remain unchanged. This isolates the effect of assuming access to the complete goal set and removes the need to propose new goals online.

\subsubsection{Known Goals and Policy Bank (BALI - KG+PB): }

This baseline is designed for the \textit{\textbf{closed case}} only, like BALI-KG, but assumes the robot has access to both a complete goal bank $\mathcal{G}$ and a policy bank $\mathcal{P}_g$ containing many ways of achieving each goal. For example, if $\mathcal{G}$ includes the recipe \textit{oatmeal}, $\mathcal{P}_g$ may contain permutations of the same preparation steps, such as \textit{get bowl} → \textit{pour oats} → \textit{add milk} → \textit{stir}, \textit{get bowl} → \textit{add milk} → \textit{pour oats} → \textit{stir}, etc. From $\mathcal{P}_g$, we derive mappings between goals to actions, and preferences to actions. These mappings support the construction of precomputed attractor fields: for each goal or preference (source), attraction scores are assigned to actions (targets) according to their frequency in associated policies. Actions consistently appearing in policies for a source receive high attraction; otherwise, they have low attraction.

Goal inference in BALI - KG+PB is handled by a Random Forest trained on 300,000 
action sequences (10,000 per recipe across 30 recipes in our specific simulation setup), 
which achieved 86.8\% accuracy on a 20\% held-out test set. During interaction, the model maps partial human action sequences to candidate goals in $\mathcal{G}$ and updates the ranking as new actions are observed. Action selection is further constrained at each node of the RHP tree by n-gram co-occurrence statistics and policy frequencies from $\mathcal{P}_g$, with candidates refined using the same LLM prompting strategy as in our method but with their costs determined entirely by precomputed attractor fields.  

\begin{figure*}[t]
    \centering
    \includegraphics[width=0.9\linewidth]{Figures/BALI_results.pdf}
    \caption{Results for the open case (a–c) and closed case (d–f). (a,d) show inference timing: maroon = time until first correct guess, yellow = period of instability until the last incorrect guess, green = stable correct phase. cross = first correct guess ($\downarrow$=better), circle = last incorrect guess($\downarrow$=better). (b,e) report Top-1 (yellow) and Top-3 (green) goal prediction accuracy ($\uparrow$=better). (c,f) show average mistakes or extra steps ($\downarrow$=better). Means and standard deviations of metrics are computed over 967 preference combinations. Visualization of standard deviations for the first correct guess and the last incorrect guess is skipped for clarity.}
    \label{fig:results}
\end{figure*}

\subsection{Metrics}

We compare BALI against each baseline using five metrics: 
\begin{enumerate}[wide, labelwidth=!, labelindent=0pt]
    \item \textbf{First Correct Guess: } The first timestep in an interaction when the robot correctly identifies the human's goal. 
    \item \textbf{Last Incorrect Guess: } The final timestep when the robot’s predicted goal is wrong, indicating when its belief stabilizes on the correct goal. This complements First Correct Guess, which may be noisy since a method can predict correctly by chance early on without resolving uncertainty. 
    \item \textbf{Goal Prediction Accuracy: } Percentage of timesteps when the robot correctly identifies the human’s goal. We report both the top-1 and top-3 accuracy. 
    \item \textbf{Number of Questions Asked:} Total clarifying questions the robot asks in an interaction.
    \item \textbf{Number of Extra Steps: } Number of extra steps taken by both agents to complete the interaction. In simulation, the metric measures extra steps in comparison to the number of ground-truth steps from the corresponding CC-HTN. Because the simulated humans make no mistakes, all extra steps are attributed to robots.
    In the real-world evaluation, we ask the human to flag any steps they think did not belong in the recipe they were trying to make.
\end{enumerate}
\section{Results in Simulation}

\subsection{Open Case}

Fig. \ref{fig:results}a–\ref{fig:results}c shows the results in simulation when the robot does not know the set of possible human goals.

\noindent
\textbf{Baselines:} Pure small LLMs made on average 13.09 mistakes per interaction, with 56.38\% top-1 and 87.30\% top-3 goal prediction accuracy. Their first correct guess occurred at 41.01\% of the interaction, and the last incorrect guess at 59.61\%, showing that predictions remained unstable well into the task. The larger LLM reduced mistakes slightly to 11.40, but top-1 accuracy dropped to 45.40\% and top-3 to 74.87\%. Their first correct guess occurred at 35.18\% of the interaction, and the last incorrect guess at 69.50\%. Our action-only baseline made 11.75 mistakes, with 49.82\% top-1 and 87.27\% top-3 accuracy, producing its first correct guess at 22.77\% and last incorrect guess at 76.67\%. GOOD \cite{ma2024goal}, which adds Bayesian reasoning over language-defined goals, achieved 56.69\% top-1 and 89.75\% top-3 accuracy, but still made 13.44 mistakes, with the first correct guess at 23.51\% and the last incorrect guess not until 59.96\%.

\noindent
\textbf{BALI:} Without asking questions, BALI reduced mistakes to 4.90, with 57.50\% top-1 and 98.09\% top-3 accuracy. The system produced its first correct guess earlier than baselines, at 20.23\% of the interaction, and its last incorrect guess by 59.80\%. With questions, performance improved further: mistakes dropped substantially to 4.10 (small model) and 3.12 (large), top-1 accuracy rose to 75.20\% and 78.90\%, and top-3 accuracy reached 98.77\% and 98.85\%. The first correct guess occurred within 10\% of the interaction, and the last incorrect guess within 6–7\%. On average, the robot asked $1.94 \pm 0.92$ questions per interaction. The LLM within BALI further improved goal prediction accuracy and reduced mistakes relative to the small model, with the gains especially pronounced in this open setting where generalization is critical.

We conducted paired t-tests across all 967 trials, confirming that all performance gaps between BALI and the baselines are statistically significant. Refer to the appendix for detailed statistics.

\begin{figure*}[t]
    \centering
    \includegraphics[width=0.78\linewidth]{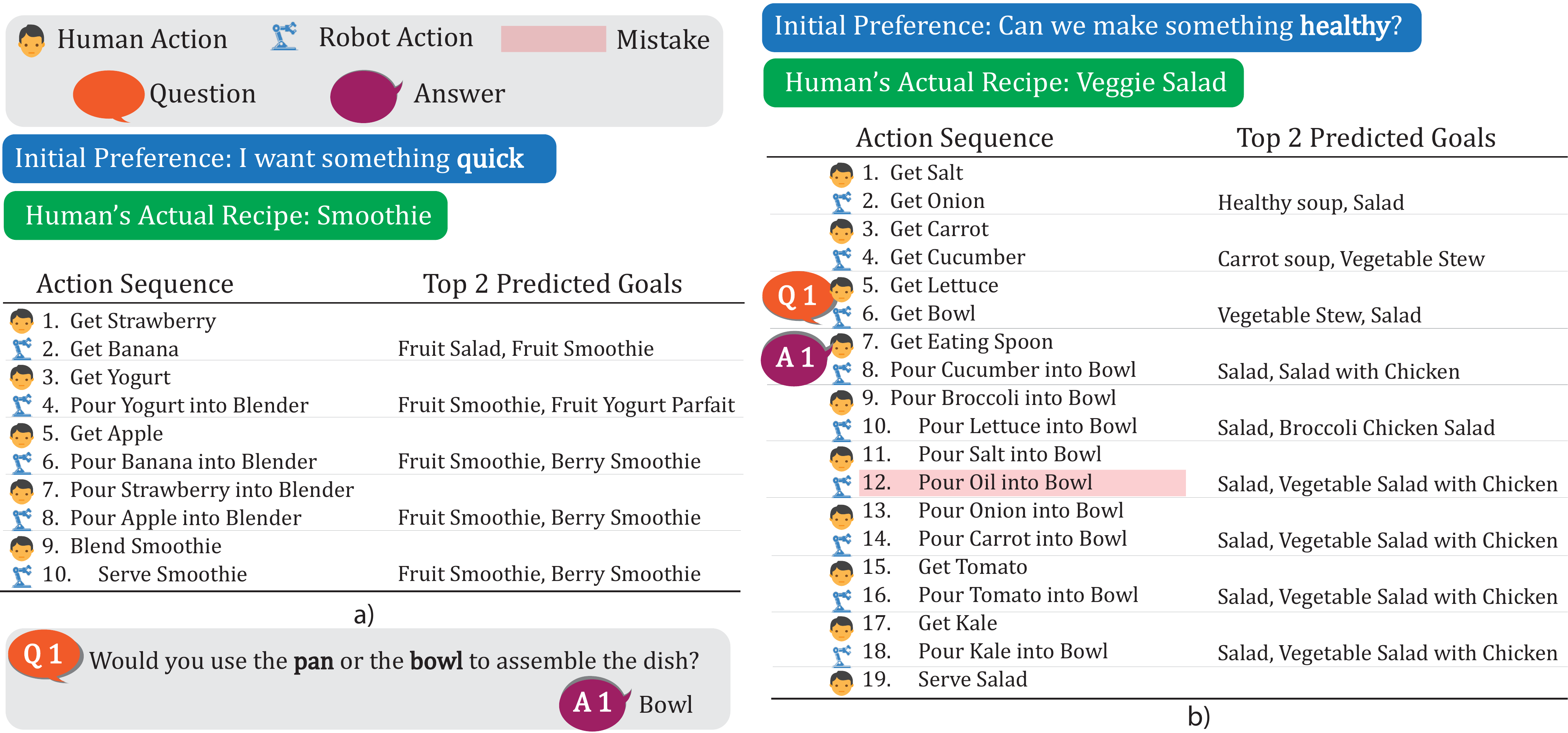}
    \caption{Real World Case Study}
    \label{fig:sequence}
    \vspace{-1em}
\end{figure*} 

\subsection{Closed Case}
Fig.~\ref{fig:results}d–\ref{fig:results}f show the results when the robot accessed the goal bank.

\noindent
\textbf{Baselines: } Similar to the open case, we see the same patterns. Pure small LLMs made 6.20 mistakes, with 53.52\% top-1 and 83.87\% top-3 goal prediction accuracy. Their first correct guess came at 26.10\% of the interaction and the last incorrect at 62.12\%, showing late convergence. Large LLMs reduced mistakes slightly to 6.21, but accuracy dropped slightly to 51.20\% top-1 and 83.06\% top-3, with first correct at 26.11\% and last incorrect at 62.80\%. GOOD \cite{ma2024goal} reached 49.05\% top-1 and 82.14\% top-3 with 4.98 mistakes, but converged late (first correct 38.32\%, last incorrect 66.17\%). Our actions-only baseline, CDP \cite{ghose2024planning} performed better than the other baselines, with 2.12 mistakes, but still performed worse than BALI.

\noindent
\textbf{BALI: } BALI-KG, the same method as in the open case but evaluated with a fixed goal set, made 1.92 mistakes, with 62.54\% top-1 and 88.92\% top-3 accuracy. The first correct guess occurred at 24.96\% and the last incorrect at 59.50\%. With questions, mistakes dropped to 1.64, with 63.98\% top-1 and 90.32\% top-3 goal prediction accuracy, first correct at 23.01\% and last incorrect at 41.68\%. Adding the policy bank (BALI-KG+PB) yielded the strongest results, as expected: mistakes nearly vanished (0.12), with 68.43\% top-1 and 93.85\% top-3 accuracy, first correct at 21.96\% and last incorrect at 57.96\%. Mistakes disappeared entirely when the robot was allowed to ask questions; top-1 accuracy reached 95.04\%, and top-3 reached 98.04\%. The first correct guess occurred at 4.90\% and the last incorrect at 9.30\%. The system asked $1.23 \pm 0.56$ questions per interaction. This shows that the question-asking module partitions the goal space effectively, so answers remove ambiguity about goals well. The large model was not tested here since the small one had already achieved near-perfect performance.

\section{Real World Demonstration}

We analyzed BALI in 3 proof-of-concept real-world trials, involving participants from the research group who were not involved in the work. The robot prompted participants to take an action to start the collaboration and provide a food preference in natural language, allowing them to add more preferences at any time. When uncertain, it asked clarifying questions to adapt based on feedback. At the end of the interaction, participants reported their goal recipe and annotated robot actions that did not align with their goal. 

Fig.~\ref{fig:sequence} shows two case studies (a third is included in the supplementary video). In the first case (Fig.~\ref{fig:sequence}a), the participant asked for \textit{“something quick”} and began preparing a smoothie. Based on this preference, the robot predicted fruit-based recipes as likely goals. It started processing ingredients early in the interaction by blending instead of continuing to fetch additional items, which the participant later noted. In the second case (Fig.~\ref{fig:sequence}b), the participant asked for \textit{“something healthy”} and began preparing a veggie salad. The vegetables fetched were consistent with multiple possible goals, so the robot asked a clarifying question about the food preparation container (“\textit{Would you use the pan or the bowl to assemble the dish?}”) to disambiguate the intended recipe. Based on the answer (“\textit{The bowl}”), it aligned its actions with their goal of making a salad, though one step of \textit{pouring oil into bowl} was flagged as a mistake. 

In real-world experiments, our implementation achieved an average inference latency of 12–14 seconds per timestep on an NVIDIA RTX 4090 GPU, using an RHP tree of depth 2 with parallelized low-latency LLM calls with the small model.

\section{Discussion}

Our results demonstrate that language and actions complement one another in revealing human goals. Explicit preferences expressed in language help interpret the intent behind actions, while observed actions clarify ambiguous or open-ended instructions. This bidirectional coupling allows the robot to refine human language with actions and ground its supportive actions in human preferences. More concretely, BALI integrates language and action in a receding-horizon planner that imagines possible futures of the collaboration, yielding earlier and more stable convergence than language-only or action-only approaches. One unique capability of our method is integrating clarifying questions into the inference loop, balancing interruption cost against expected information gain. This allows the robot to resolve ambiguity efficiently without burdening the human by asking for instructions to improve collaboration fluency. 

Our work bridges two paradigms of goal inference that have traditionally been studied in isolation: closed-set recognition, where robots classify among a finite set of goals, and open-ended reasoning, where goals may be unbounded and novel. Research on goal prediction has primarily remained in the closed case \cite{ghose2023tailoring, ghose2024planning, ghose2025adapting, wu2021too, jain2019probabilistic}, with the open case largely avoided by constraining the problem to predefined goals. The few open-ended approaches, such as GOOD \cite{ma2024goal}, rely exclusively on dialogue. By contrast, BALI fuses observed actions with preferences, grounding open-ended goals in the collaboration where intent is expressed through behavior and language.

Our work has limitations, which are opportunities for future work. First, the experiments assumed fixed turn-taking to induce equal participation between the human and the robot, but this is not a fundamental requirement of our method. Second, BALI was evaluated 
assuming the human's end goal remains fixed throughout the interaction. 
Evaluating BALI when human goals change over time or when human behavior is suboptimal (like \cite{ghose2025ve}) would be interesting. Third, in real-world deployments, BALI can incur noticeable latency from LLM-based scoring and tree search; this could be reduced by more aggressive pruning of candidate actions, caching attractor-field computations, or using smaller or quantized fine-tuned LLMs. Finally, BALI relies on LLMs as judges to score attractor fields in the open case, introducing variability and opacity. Future work includes addressing these issues through more interpretable scoring methods, studying human perceptions of clarifying questions in real deployments, and extending BALI with algorithms for open-world planning and goal reasoning \cite{chen2025planning, athalye2024pixels}.

\section*{Acknowledgments}

This work was supported by the National Science Foundation (NSF) awards  IIS-2143109, IIS-2106690, and IIS-1955653, and Office of Naval Research (ONR) award N00014-24-1-2124. We thank Tesca Fitzgerald, Kate Candon, Nicholas Georgiou, Jirachaya ``Fern" Limprayoon, and Shasvat Desai for help in improving this work. 

\balance

\bibliographystyle{ACM-Reference-Format}
\bibliography{main}


\end{document}